\documentclass[sigconf]{acmart}

\AtBeginDocument{%
  \providecommand\BibTeX{{%
    \normalfont B\kern-0.5em{\scshape i\kern-0.25em b}\kern-0.8em\TeX}}}

\copyrightyear{2023}
\acmYear{2023}
\setcopyright{acmlicensed}\acmConference[KDD '23]{Proceedings of the 29th ACM SIGKDD Conference on Knowledge Discovery and Data Mining}{August 6--10, 2023}{Long Beach, CA, USA}
\acmBooktitle{Proceedings of the 29th ACM SIGKDD Conference on Knowledge Discovery and Data Mining (KDD '23), August 6--10, 2023, Long Beach, CA, USA}
\acmPrice{15.00}
\acmISBN{979-8-4007-0103-0/23/08}
\acmDOI{10.1145/3580305.3599246}

\usepackage{multirow}
\usepackage{tabu}
\usepackage{balance}

\newcommand{\para}[1]{\noindent\textbf{#1.}}

\begin{document}

\title{A Study of Situational Reasoning for Traffic Understanding}

\author{Jiarui Zhang}
\email{jrzhang@isi.edu}
\affiliation{%
  \institution{Information Science Institute, University of Southern California}
  \city{Los Angeles}
  \state{California}
  \country{USA}
}

\author{Filip Ilievski}
\email{ilievski@isi.edu}
\affiliation{%
  \institution{Information Science Institute, University of Southern California}
  \city{Los Angeles}
  \state{California}
  \country{USA}
}

\author{Kaixin Ma}
\email{kaixinm@cs.cmu.edu}
\affiliation{%
  \institution{Language Technologies Institute, Carnegie Mellon University}
  \city{Pittsburgh}
  \state{Pennsylvania}
  \country{USA}
}

\author{Aravinda Kollaa}
\email{kollaa@isi.edu}
\affiliation{%
  \institution{Information Science Institute, University of Southern California}
  \city{Los Angeles}
  \state{California}
  \country{USA}
}

\author{Jonathan Francis}
\email{jon.francis@us.bosch.com}
\affiliation{%
  \institution{Human-Machine Collaboration, Bosch Center for Artificial Intelligence}
  \city{Pittsburgh}
  \state{Pennsylvania}
  \country{USA}
}

\author{Alessandro Oltramari}
\email{alessandro.oltramari@us.bosch.com}
\affiliation{%
  \institution{Human-Machine Collaboration, Bosch Center for Artificial Intelligence}
  \city{Pittsburgh}
  \state{Pennsylvania}
  \country{USA}
}

\renewcommand{\shortauthors}{Jiarui Zhang, et al.}

\begin{abstract}
Intelligent Traffic Monitoring (ITMo) technologies hold the potential for improving road safety/security and for enabling smart city infrastructure. Understanding traffic situations requires a complex fusion of perceptual information with domain-specific and causal commonsense knowledge. Whereas prior work has provided benchmarks and methods for traffic monitoring, it remains unclear whether models can effectively align these information sources and reason in \textit{novel} scenarios. To address this assessment gap, we devise three novel text-based tasks for situational reasoning in the traffic domain: i) BDD-QA, which evaluates the ability of Language Models (LMs) to perform situational decision-making, ii) TV-QA, which assesses LMs' abilities to reason about complex event causality, and iii) HDT-QA, which evaluates the ability of models to solve human driving exams. 
We adopt four knowledge-enhanced methods that have shown generalization capability across language reasoning tasks in prior work, based on natural language inference, commonsense knowledge-graph self-supervision, multi-QA joint training, and dense retrieval of domain information. We associate each method with a relevant knowledge source, including knowledge graphs, relevant benchmarks, and driving manuals.
In extensive experiments, we benchmark various knowledge-aware methods against the three datasets, under zero-shot evaluation; we provide in-depth analyses of model performance on data partitions and examine model predictions categorically, to yield useful insights on traffic understanding, given different background knowledge and reasoning strategies.
We have made our code and data publicly available. \footnote{\url{https://github.com/saccharomycetes/text-based-traffic-understanding}}

\end{abstract}

\ccsdesc[500]{Computing methodologies~Natural language processing}
\ccsdesc[300]{Computing methodologies~Knowledge representation and reasoning}
\ccsdesc[300]{Computing methodologies~Learning paradigms}


\keywords{traffic understanding, zero-shot evaluation, language models, question answering}

\begin{teaserfigure}
\centering
  \includegraphics[width=0.7\textwidth]{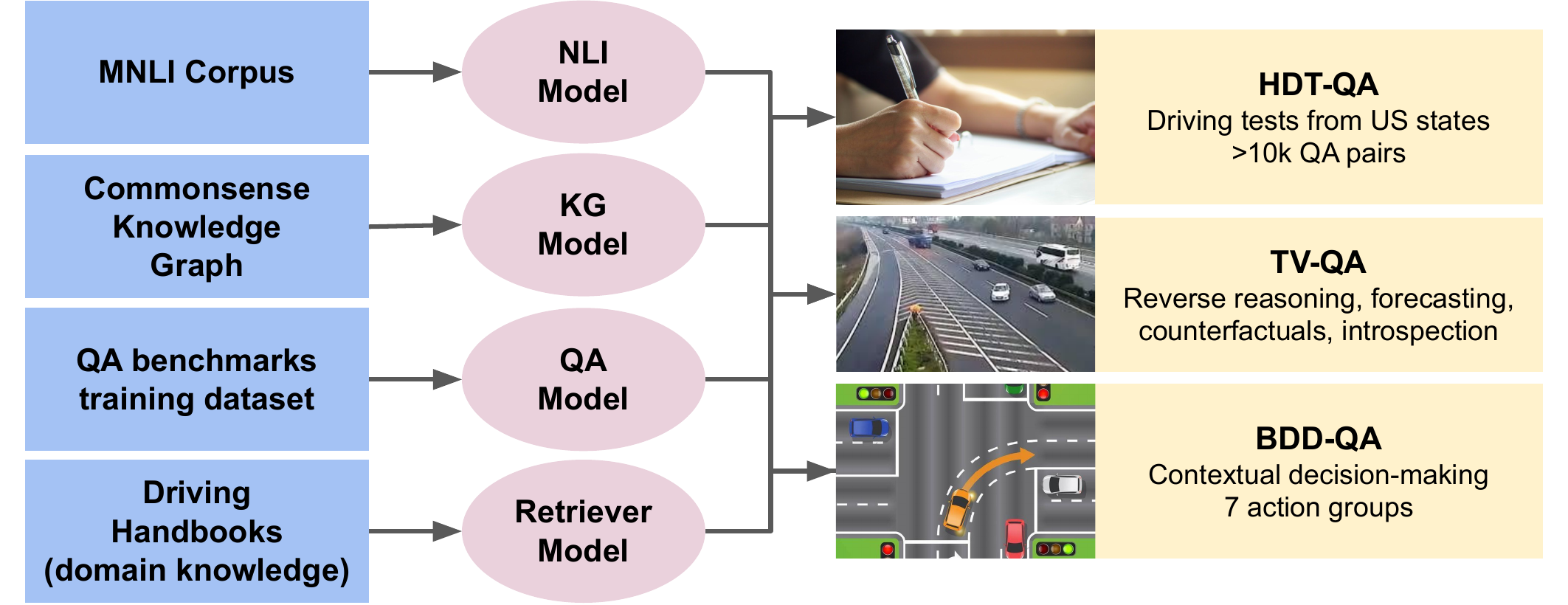}
  \caption{Overview of our study framework, which evaluates four knowledge-enhanced language methods, adapted with different knowledge sources on three traffic domain datasets, in a zero-shot manner.}
  \label{fig:teaser}
\end{teaserfigure}


\maketitle

\section{Introduction}



As humans, we are able to make sense of novel situations even with limited domain-specific knowledge. For machine reasoning, situational understanding has recently seen a resurgence of commonsense reasoning technologies, demonstrated in tasks such as question answering \cite{ma2021exploring,zhang2022empirical} and story comprehension \cite{ma2022coalescing, Cui_Che_Zhang_Liu_Wang_Hu_2020}. However, the ability of models to generalize to novel situations in specific domains has been relatively understudied.
The domain of Intelligent Traffic Monitoring (ITMo), which represents an essential instrument to improve road safety and security, is of particular interest as it supports the vision of developing smart city infrastructures of the future~\cite{chowdhury2021towards}. ITMo has accumulated a significant market value from car manufacturers~\cite{IntelligentTraffic}, as well as significant government support, e.g., through a recent infrastructure bill~\cite{qasemi2022intelligent}. 


Being able to understand novel traffic situations requires a complex association of perceptual information with assumed domain-specific and causal commonsense knowledge. For example, reasoning whether an accident could be prevented by slower movement or forecasting the consequence of different movement patterns of a vehicle depends on the specific situational circumstances (e.g., the velocity of the other vehicles, road conditions), domain reasoning (e.g., what is the maximum allowed speed on a freeway), and causal reasoning (e.g., what could happen if the car changes lanes immediately, or what can be expected behavior from the other traffic participants). While prior work has developed evaluation tasks for traffic monitoring, only a few of them are available to the public~\cite{xu2017end, kim2018textual,xu2021sutd}. These tasks have inspired corresponding methods~\cite{halilaj2021knowledge,muppalla2017knowledge,chowdhury2021towards,wickramarachchi2020evaluation}, which usually focus on associating perception signals, e.g., from video cameras, with answers to natural language questions. While providing an important step in the right direction, these tasks and methods do not provide mechanisms for studying models' holistic reasoning abilities over novel situations in traffic. 

Given that language is a natural medium for human communication, it is intuitive to use the textual modality to develop and test robust traffic understanding models.
A large body of work has focused on general commonsense reasoning~\cite{ma2019towards, ma2021exploring, zhang2022empirical}, but the role of commonsense and causal reasoning has not been explored thoroughly in the traffic domain. Complementarily, ontologies for describing traffic participants~\cite{OpenXOntology} exist, but it is unclear whether connecting them to neural models can bring the desired generalizability in reasoning. 
To the best of our knowledge, understanding traffic situations presented in natural language has not been explored in depth so far. This brings up a natural question of whether large pretrained models can effectively solve realistic traffic understanding tasks dominant in natural language, such as decision-making in the context of a given situation, reasoning about the causes and effects of a given (real or hypothetical) situation, and answering questions that rely on domain expert knowledge. Moreover, as state-of-the-art traffic monitoring methods are often designed as black boxes, prior work has not %
systematically investigated the role of various background knowledge types (e.g., causal commonsense knowledge and traffic domain knowledge) in reasoning over traffic situations. 



In this paper, we \textit{study the ability of diverse knowledge-enhanced language models to reason over traffic situations}. By doing so, we are able to reduce the complexity of traffic reasoning to a single modality (text), and measure the success of different method families and knowledge sources on representative traffic monitoring tasks. %
A schematic overview of our study framework is shown in Figure~\ref{fig:teaser}. %

We summarize the contributions in this paper as follows:

\noindent (1) We devise \textbf{three novel text-based tasks for situational reasoning in the traffic domain}: \textit{i) BDD-QA}, which evaluates the ability of LMs to perform situational decision-making, \textit{ii) TV-QA}, which assesses LM's ability to reason about complex event causality, and \textit{iii) HDT-QA}, which evaluates the ability of models to solve human driving tests. We divide each of these datasets into meaningful partitions to support fine-grained analysis. \\
\noindent (2) We adopt \textbf{four knowledge-enhanced methods that have been shown to generalize well} across natural language reasoning tasks in prior work, based on natural language inference, commonsense knowledge graph (KG) self-supervision, multi-task QA joint training, and retrieval-augmented QA. We associate each of these methods with a relevant knowledge source, including knowledge graphs, relevant benchmarks, and driving manuals.\\
\noindent (3) We perform \textbf{extensive experiments of different knowledge-aware methods against the three datasets in a zero-shot evaluation setup}. We perform an in-depth analysis of model performance on data partitions and look closer into the model predictions to provide useful insights into the traffic understanding ability of different background knowledge and reasoning methods.

\section{Related work}


\subsection{Traffic Understanding}

CADP~\cite{shah2018cadp} is a spatiotemporally annotated dataset for accident forecasting using traffic camera views.
The Berkeley Deep-Drive dataset (BDD) is a
video dataset consisting of real driving videos containing abundant driving scenarios~\cite{xu2017end}. The follow-up work, BDD-X \cite{kim2018textual}, provides the action description of the BDD video and their explanations. 
TrafficQA \cite{xu2021sutd} consists of over 60K QA samples based on over 10k traffic scenes. The QA set of TrafficQA includes 6 different aspects of reasoning problems: basic understanding, attribution, introspection, counterfactual inference, event forecasting, and reverse reasoning. All QA pairs are based on visual inputs. The traffic benchmarks have inspired corresponding methods: by using knowledge to enhance traffic understanding,~\citet{wickramarachchi2020evaluation} create and evaluate knowledge graph embeddings for autonomous driving. \citet{chowdhury2021towards} enhance the knowledge graph for scene entity prediction in autonomous driving. ITSKG~\cite{muppalla2017knowledge} is a knowledge graph framework for extracting actionable information from raw sensor data in traffic. CoSI~\cite{halilaj2021knowledge} is a knowledge graph-based approach for representing information sources relevant to traffic situations. 
Different from the dominant traffic monitoring focus on the perception modality, we explore a shift in paradigm, from monitoring to \textit{understanding}, which requires abundant domain knowledge and commonsense reasoning methods. However, perceptual annotations provide valuable information that can be reused for creating comprehension tasks. We utilize textual descriptions of actions and their explanations associated with the BDD-X dataset to construct our BDD-QA dataset for causal reasoning. Similarly, we leverage the TrafficQA data to create a text-based task by transcribing the videos into textual descriptions.

\subsection{Situational Reasoning in Natural Language}

MultiNLI (MNLI) \cite{williams2017broad} is a large (433k) corpus for the natural language inference (NLI) task---with ten distinct genres of written and spoken English. MNLI has become a popular benchmark for assessing various language models, such as BERT~\cite{devlin2018bert}, RoBERTa~\cite{roberta}, BART~\cite{bart}, and DeBERTa~\cite{deberta}. Although they are widely used in general reasoning tasks, these methods have not been applied to the traffic domain, to the best of our knowledge. In our work, we use such models that are pre-trained on MNLI to evaluate their zero-shot performance on inference tasks in the traffic domain, based on the assumption that the models are generally capable of performing causal reasoning. ATOMIC \cite{sap2019atomic} provides a large atlas of everyday commonsense knowledge, focusing on inferential knowledge of causes and effects (i.e., if-then-else conditions). CauseNet \cite{heindorf2020causenet} provides a large-scale knowledge base of claimed causal relations between concepts which can be used for casual reasoning and question-answering tasks. \citet{Ma2021} developed a method that utilizes different types of commonsense knowledge, extracted from a federated knowledge graph, to train language models via self-supervision: they found that such models yield high accuracy across several natural language tasks, under zero-shot evaluation~\cite{zhang2022empirical}. Despite knowledge-driven models performing well on many general-domain commonsense reasoning tasks, the question remains how to customise these approaches for a specific domain. Therefore, we assess performance of modeling approaches on reasoning tasks in the traffic domain, to investigate the effect of commonsense knowledge on answering domain-specic questions. 
\section{Tasks \& Datasets}

In this section, we describe the three diverse datasets we construct for comprehensively evaluating models' diverse situational reasoning strategies in the traffic domain---BDD-QA, TV-QA, and HDT-QA. We refer to situational reasoning as the task of providing the best judgment or decision for a given situation. 
Formally, we pose traffic situational reasoning as a multiple-choice question-answering task. Each task entry consists of a natural language question $Q$ and $n$ candidate answers $\mathbf{A} = \{A_{1}, ..., A_{n}\}$. 
Given $Q$ and $\mathbf{A}$, a model's prediction task is to select the single best answer from the candidates.
%
We show statistics and examples from our data in tables \ref{tab: statistic} and \ref{tab: qaexaples}.

\begin{figure}[ht]
\centering
\includegraphics[width=1\linewidth]{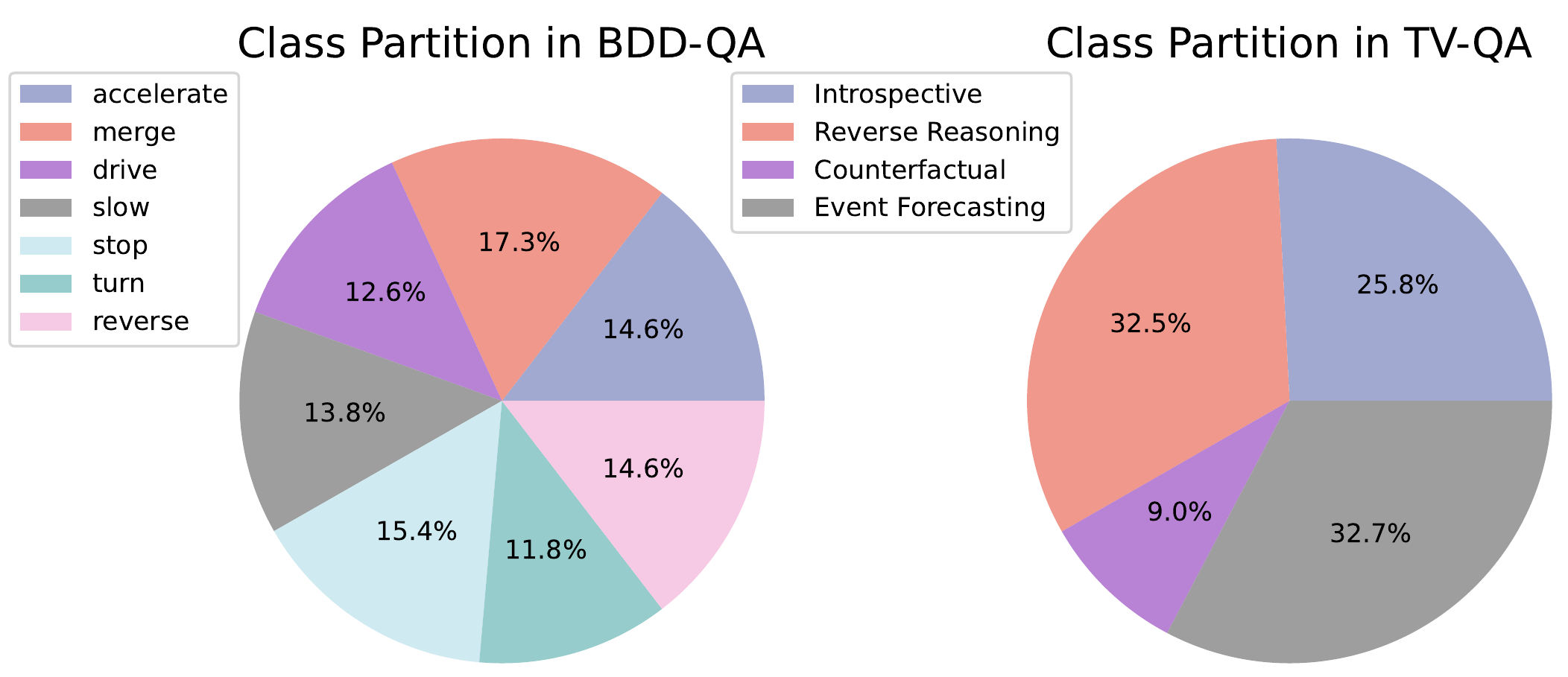}\\
\caption{BDD-QA and TV-QA data partitions.} 
\label{fig:partition}
\end{figure}

\subsection{BDD-QA}

The first dataset we construct is BDD-QA, which tests models' decision-making capabilities by asking for the best action in a given situation from a driver's perspective.

\para{Data collection} We create BDD-QA from the Berkeley Deep-Drive Explanation Dataset \cite{kim2018textual} (BDD-X). The dataset consists of human annotations on various traffic videos covering diverse weather conditions and road types. In total, BDD-X provides over 26K sentence pairs where each pair consists of a description of an action in the video and its justification (reason). 

\para{Task preparation} The \{\texttt{action}, \texttt{justification}\} pairs naturally function as questions and answers. To convert them into multiple-choice questions, a key challenge is an effective sampling of distractors that are both fair (i.e., incorrect answers) and informative (i.e., sufficiently relevant to support a meaningful task). Similar to prior work~\cite{Ma2021}, we observe that random sampling alone would result in a low-quality dataset because the distractors might also be plausible answers. Upon both the automatic analysis of the data by sentence transformers~\cite{reimers2019sentencebert} and k-means clustering, as well as manual analysis of 100 data samples, we noted that the actions covered in this data are relatively limited and they can usually be associated with the sentence verbs. 
Inspired by this observation, we extract the verbs using Spacy~\cite{spacy2} 
and count the frequency of the verbs in the whole set. We pick the verbs with a frequency higher than 20 and manually filter the incorrectly recognized verbs. 
We manually distill
7 classes from the remaining set of verbs to represent the car's actions, namely: \texttt{accelerate}, \texttt{move}, \texttt{slow}, \texttt{stop}, \texttt{merge}, \texttt{turn}, and \texttt{reverse}. 
We apply these actions to classify the questions based on their verbs. 
We filter all of the repetitive sentence pairs and remove the actions with multiple verbs to make sure the actions can be correctly classified.

After the verbs are classified into actions, we sample three distractors from different action classes for each question, which leads to 14,039 multiple-choice questions in total. However, a rule-based sampling strategy cannot ensure data quality, and semantically repetitive questions and multiple reasonable answers may still exist.
To obtain high-quality test data, we randomly sample 50 examples from every class. We then conduct a thorough manual refinement process to replace possibly unfair distractors with admissible ones, hence only one answer is plausible. We also remove semantically repetitive questions. The resulting 255 multiple-choice questions are used for evaluation. The partitions of the test dataset are presented in Figure~\ref{fig:partition}, showing a relatively balanced distribution among the seven action classes.

\subsection{TV-QA}
\label{ssec:tv-qa}
The second dataset we constructed is TV-QA, which evaluates models' capabilities to perform complex causal analyses of real and hypothetical traffic events from an observer's perspective.

\para{Data collection} We build TV-QA from the recently proposed Traffic-QA dataset \cite{xu2021sutd}. The questions in Traffic-QA are based on corresponding videos, recording different types of traffic events (accidents, vehicle turning, pedestrian behaviors, etc.). The videos are 7 seconds long on average, and each video is associated with one or multiple questions. Traffic-QA categorizes questions into 6 traffic-related reasoning types: basic understanding, attribution, event forecasting, reverse reasoning, counterfactual inference, and introspection. 
Among the reasoning types, the "basic understanding" category is the most populated and it involves asking questions about the basic facts in the video, such as "Is there a crossroad in the video?", which is more akin to a computer vision recognition task rather than our focus of study. We also observe similar cases in the "attribution" category. 
As a result, we select questions from the latter four reasoning types to be included in our dataset. Two of these, "Event Forecasting" and "Reverse Reasoning", are surface-level natural language reasoning tasks that involve predicting events based on descriptions of previous and following events. The remaining two tasks, asking "What could happen if..." (Counterfactual reasoning) and "What should have been done to..." (Introspection), require an understanding of the causes and effects of the entire sequence of the traffic events and their alternatives, as well as a high degree of experience to provide reasonable predictions.

\para{Task preparation} To gather detailed human descriptions of the videos, we established a crowd-sourcing task with Amazon Mechanical Turk to elicit descriptions from annotators. To maximize the level of detail in the descriptions, we provided annotators with three questions pertaining to the initial scene~(1 sentence), the procedure~(2-3 sentences), and the final scene~(1 sentence) of each video. We selected 200 videos for this annotation and asked three annotators to provide descriptions for each video.
After obtaining the descriptions, to ensure that they contain sufficient information to support question answering about the traffic event, we set another crowd-sourcing task in which we show the annotators the description, question, and candidate answers. This task included 783 questions corresponding to the original 200 videos, and 3 annotators answer each description-question-answers triple. After collecting the answers given by the annotators, we choose the questions that are correctly answered by the majority of the annotators (at least 2 out of 3) using at least one description of each video as our valid QA dataset. This ensures that the descriptions given by annotators are sufficient for answering the questions; this results in 434 questions and 111 videos in total.
Finally, we manually revise the descriptions of each valid description-question-answer triple and use the validated data as our test set. The partition of each reasoning task in the test dataset is presented in Figure~\ref{fig:partition}. We note that all reasoning types have a similar participation in TV-QA, except counterfactual inference, which has a much smaller set of examples.

\begin{table}[!t]
	\centering
	\small
	\caption{Statistics of the datasets by train/test split. }
	\label{tab: statistic}
	\begin{tabular}{l| ccc }
	\hline
        Dataset\&Partitions&{\bf{BDD-QA}}&{\bf{TV-QA}}&{\bf{HDT-QA}}\\\hline
        \bf Train&14,039&-&9,456\\
        \bf Test &255&434&500\\\hline
	\end{tabular}
\end{table}

\begin{table}
\centering
\small
\caption{Examples of BDD-QA, TV-QA, and HDT-QA. (*) denotes the correct answer.}
\label{tab: qaexaples}
\begin{tabular}{p{8.5cm}}
\tabucline[1.1pt]\\
\textbf{BDD-QA} \\
\textit{Q:} The car in front of the car is slow, \\
but the traffic is also heavy in other lanes, what will the car do next? \\
\textit{A1:} The car speeds up and turns to the right; \\
\textit{A2:} The car moves back to the right side of the road; \\
\textit{A3:} The car slows down(*);\\
\textit{A4:} The car backs up slowly\\\tabucline[1.1pt]\\
\textbf{TV-QA} \\
\textit{Description:} The POV car is quickly going down a highway. 
The POV car approaches an intersection.
There is a red sedan in the opposing lane
waiting to turn and cross the intersection.
The red sedan quickly makes a left turn.
when the POV car enters the intersection. 
The POV car veers to the right. The red sedan hits the side of the POV car.\\
\textit{Q}: Could the accident be prevented if the involved vehicles change lane or turn properly?\\
\textit{A1:} Yes(*);\\
\textit{A2:} No, that was not the main cause of the accident\\\tabucline[1.1pt]\\
\textbf{HDT-QA} \\
\textit{Q:} If you find yourself in a skid:\\
\textit{A1:} Brake lightly;\\
\textit{A2:} Brake abruptly;\\
\textit{A3:} Stay off the brakes(*)\\\tabucline[1.1pt]\\
\end{tabular}
\label{tab:examples}
\end{table}


\subsection{HDT-QA}
\label{ssec:hdtqa}
The third dataset we construct is HDT-QA, which focuses on traffic domain knowledge, including driving policies and driver decisions.

\para{Data collection} Human driving tests provide a natural testbed for evaluating domain knowledge about traffic. Sample tests publicly available on the Web can be used to help people assess their knowledge, but also to evaluate the ability of machines to do the same. We scrape the \textbf{H}uman \textbf{D}riving \textbf{T}est \textbf{Q}uestion \textbf{A}nswering (HDT-QA) dataset from a website that contains 51 collections of sample driving tests pertaining to the 50 USA states and Washington D.C.~\cite{Drivingtest}. For each state collection, there are three subjects of driving tests and manuals: \textit{Motorcycle}, \textit{Car} and \textit{Commercial
Driver}---resulting in 153 state-subject combinations. Each test consists of set of multiple-choice questions, where the number of answer candidates ranges between 3 and 4. 

\para{Task preparation} As the driving tests are already comprised of multiple-choice questions, we can use these questions to evaluate domain knowledge comprehension by AI models.
To make the test data fair for text-based methods, we only keep questions that can fully be answered based on the textual content, i.e., we exclude all questions with images (e.g., traffic signs). We also filter out the questions where one of the candidate answers is \textit{all of the above}, as answering such questions is not the focus of our work. We amalgamate the test datasets from all states and subjects, eliminating duplicated questions and consolidating them based on the number of candidate answers. We split the resulting data into 9,456 train questions and 500 test questions.

\section{Method}
\label{sec:methods}

While fine-tuned language models achieved state-of-the-art performance on many language-based reasoning tasks~\cite{devlin2018bert,radford2019language}, prior work has shown that this could lead to over-fitting~\cite{ma2021exploring}. Anticipating similar over-fitting to QA sets in the traffic domain, we select representative existing methods that utilize background knowledge to perform reasoning on unseen QA tasks in different ways.
We use four methods that enrich language models with background knowledge relevant for the traffic domain, based on: Natural Language Inference (NLI), Knowledge Graph self-supervision, Question Answering, and Retrieval-augmented QA. 



\subsection{Natural Language Inference}

The task of NLI requires the model to select the most plausible candidate from a set of hypotheses given a premise. A multiple-choice QA task can be seen as a union of several NLI tasks, where each corresponds to performing inference over one of the choices.
To reduce any potential training/test format mismatch, we keep our QA data as declarative sentence pairs.
The two sentences are given to an NLI model as the input. 
Given a sentence pair $\{S_1, S_2\}$, the model is asked to respectively score entailment, neutral, and contradiction, for each sentence, as $\{P_e, P_n, P_c\}$. 
Then, the model chooses the candidate with the highest margin score of $P_e-P_c$ as the true answer.
Formally, the final prediction of the model will be:
$$
pred=
\arg \max_{i}(P_e^i - P_c^i),
$$
where $i$ denotes the $i_{th}$ candidate sentence pair for the model to choose. 
%
%
%
For our experiments, we adopt an off-the-shelf NLI model and apply it directly to our three datasets. In particular, we select the RoBERTa model fine-tuned on the MultiNLI~\cite{williams2017broad} dataset, which is one of the largest training corpora for NLI, consisting of 433k data entries covering ten genres.  

\subsection{Knowledge Graph Self-Supervision}
\label{sec: kgmethod}

Self-supervised training with structured knowledge has shown promising results in recent works~\cite{Ma2021,zhang2022empirical,dou2022zero}. These methods automatically construct synthetic data from commonsense knowledge graphs and then train the model on the synthetic data. Due to the broad coverage of knowledge in diverse knowledge graphs, ~\cite{ilievski2021cskg}, models trained on synthetic data show strong generalization abilities to unseen commonsense reasoning tasks. Thus, we investigate whether these models can generalize well to the traffic domain too. We adopt the RoBERTa-large model and T5-large model released by ~\citet{zhang2022empirical}, as well as their corresponding scoring methods. 

Given a question and several candidates, the model chooses the question-candidate pair that best entails the knowledge extracted from the knowledge graph:

$$
pred=\arg \max_{i}(R(s_{i}|KG))
$$
where $R(s|KG)$ is the reasonableness score of a sentence based on the knowledge that the language model extracted from the knowledge graph.

The RoBERTa model uses the Mask Language model Loss to give a score for the combination of question and candidate pairs. For the T5 model, the prefix "reasoning" is added at the beginning of each question-candidate pair. We define the loss of T5 models~(sequence to sequence loss) generating a string $St$ as $L_{St}$
then the reasonableness score is $S=L_{"1"}-L_{"2"}$ because the model is trained to generate a "1" token for reasonable input and a "2" for unreasonable input.
Both of these models 
are pretrained on a synthetic dataset with 1 million short question-answer pairs generated from the Commonsense Knowledge Graph~\cite{ilievski2021cskg}. 

\subsection{Question Answering}
\label{ssec:qa}
Recent work shows that training a multi-task QA model on a diverse range of QA datasets achieves strong generalization ability on unseen datasets~\cite{khashabi2020unifiedqa}.
Thus, we adopted such a multi-task QA model on our traffic domain benchmarks. Specifically, we used the Unified-QA-v2 models released by~\citet{khashabi2022unifiedqa}, which is trained on 20 QA benchmarks crossing different domains and question formats. Since Unified-QA converts all tasks into a sequence-to-sequence format, we also follow this strategy in our evaluations. 
Given a question $Q$, and a set of answer candidates $\{A_1, A_2,...\}$, we concatenate the question with all answer candidates as the input. Then the model is expected to generate the correct answer as output. We calculate the difference~\footnote{We calculate the difference between the output sequences by finding the longest contiguous matching sub-sequences.} between the T5 output and each candidate, and we choose the candidate with the least difference as the choice of the model.





\subsection{Retrieval-Augmented QA}
The aforementioned methods all make the assumption that the knowledge required to predict the answer is already learned by the models, i.e., encoded in their parameters as implicit knowledge. However, encoding the vast amount of knowledge in the model parameters may be unrealistic for many knowledge-intensive tasks \cite{lewis2020retrievalaugmented}. As an alternative, augmenting the QA model with relevant information retrieved on the fly shows promising results in many recent studies \cite{ma-etal-2022-open,mcdonald2022detect}. Thus, we also adopted a retrieval-augmented approach in our evaluation. In particular, we adopted the Dense Passage Retrieval (DPR)~\cite{dpr} to find relevant domain knowledge.
DPR is a neural retriever model trained on open-domain QA datasets that require Wikipedia knowledge \cite{gnq}. The DPR model is shown to outperform traditional lexical-based retrieval methods like BM25 on those popular benchmarks. 

In our case, we use DPR to encode our traffic domain corpus (described next) and retrieve relevant documents for each question. Given a corpus $C \in \{p_1, p_2, ...p_n\}$ and a question Q, where $p_i$ are individual passages. We use the dot-product score to extract the most relevant paragraph:
$$
p_{select} = \arg \max_{i}~dot(\hat{p_i},\hat{Q}), p_i \in C
$$
where $\hat{p_i}$ and $\hat{Q}$ are the passage and the question respectively as encoded by DPR. To perform retrieval-augmented QA, we feed the top-1 retrieved passage along with the question and its answer candidates to the Unified-QA model to get the final answer. Here, we simply concatenate the retrieved to the front of the question and then run inference in the same way as described in \autoref{ssec:qa}. 

\para{Retrieval corpus} We scrape driving manuals from all 50 USA states from a comprehensive website~\cite{Drivingtest}. Analogous to the human driving tests (cf. \autoref{ssec:hdtqa}), for each state, there are three subjects of driving manuals (\textit{Motorcycle}, \textit{Car}, and \textit{Commercial Driver}), resulting in 150 state-subject combinations. 
Within our study, we treat the comprehensive collection of driving manuals as a domain corpus~\cite{Drivingtest}. After processing the manual documents, we divide them into sentences by simply using the punctuation \textit{" . ? ! "} as sentence boundaries. As information extraction from PDF is imperfect, we use a T5-based grammar correction \cite{GrammarCorrection} as a data-cleaning tool to improve the spelling and grammar of the corpus. Finally, we combine every 10 consecutive sentences as paragraphs that are fed to the QA model.\footnote{We also attempted to segment the sentences into paragraphs based on the similarity between neighboring sentences, but this strategy yielded consistently worse results.} In total, our corpus consists of $42,979$ paragraphs, where each paragraph has an average token count of $147.5$.

\section{Experimental Setup}

\para{Baselines} To indicate the upper bound of models' performance on our dataset, we include the results for a supervised method based on fine-tuning of the Roberta-large model. We only do this in the BDD-QA dataset and HDT-QA dataset due to the limited size and the lack of training data of the TV-QA dataset. For lower bound performance, we include results for a random baseline and an unsupervised vanilla Roberta-large model.

\para{Evaluation}~~Our evaluation primarily involves testing all methods on all datasets in a zero-shot manner. We also perform a transfer learning experiment, where we take a supervised model trained on one dataset and evaluate it directly on another dataset. We use accuracy as an evaluation metric, calculated as the ratio of correctly answered questions to the total number of questions. 

\para{Implementation}~~Our implementation is based on python 3.7.10, PyTorch 1.9.0, and transformers 4.11.3.
For the transfer learning experiments, we use a learning rate of $1e^{-5}$, batch size of 32, weight decay 0.01, training epochs of 10, adam-epsilon of $1e^{-6}$, $\beta1=0.9,\beta2=0.98$, warm-up proportion of 0.05, and margin of 1.0.\\
\para{Computation}~~For CPU computation, we use Intel(R) Xeon(R) Gold 5217 CPU @ 3.00GHz (32 CPUs, 8 cores per socket, 263GB RAM).
For GPU, we use an NVIDIA RTX A5000.

\begin{table}[!t]
	\centering
	\small
	\caption{Evaluation results on our three benchmarks, with best results in \textbf{bold} and second best in \underline{underline} font. For BDD human evaluation, we randomly selected 100 questions from the test set, and we ask 3 humans to give their answers. We recognize the questions that are correctly answered by the majority voting~(at least 2 out of 3) as correctly answered questions. For TV-QA, the human accuracy is 100\% since the dataset is comprised of the questions that can be correctly answered by the majority voting. While we did not evaluate human performance on HDT-QA, we expect that this data is of high quality as its questions are used to evaluate human command of traffic rules and policies in the real world.}
	\label{tab: all result}
	\begin{tabular}{l | rrr}
	\hline
	    \multirow{2}{*}{\bf Method}& \multicolumn{3}{c}{\bf Benchmark}\\& \textbf{BDD-QA}&~~\textbf{TV-QA}~~&~~\textbf{HDT-QA}~~~\\\hline
        \bf Random baseline&0.250&0.373&0.268\\
        \bf RoBERTa-large~(unsupervised)&0.309&0.412&0.334\\\hline
        \bf NLI-RoBERTa-large &\underline{0.471}&\underline{0.479}&0.324\\\hline
        \bf KG-RoBERTa-large&0.443&0.465&0.352\\
        \bf KG-T5-large&0.458&\textbf{0.506}&0.344\\\hline
        \bf QA-T5-large&\underline{0.471}&0.468&\underline{0.414}\\\hline
        \bf Retrieval-T5-large&\textbf{0.537}&0.444&\textbf{0.440}\\\hline
        \bf RoBERTa-large~(supervised)&0.894&-&0.822\\
        \bf Human~&0.950&1.000&-\\\hline
	\end{tabular}
\end{table}

\begin{table*}[!t]
	\centering
	\small
	\caption{Results by action class on the BDD-QA dataset. 
 }
	\label{tab: bdd result}
	\begin{tabular}{l | rrrrrrr | r}
	\hline
	    \multirow{2}{*}{\bf Methods}& \multicolumn{8}{c}{\bf Action Class}\\
        & Accelerate&~~Move~~& ~~Slow~~~& ~~Stop~~~& ~~Merge~~& ~~Turn~~~&~~Reverse~~&~~Average\\\hline
        \bf Random baseline&0.250&0.250&0.250&0.250&0.250&0.250&0.250&0.250\\
        \bf RoBERTa-large (unsupervised)&0.405&0.312&0.285&0.051&0.477&0.500&0.162&0.309\\\hline
        \bf NLI-RoBERTa-large &0.459&\textbf{0.781}&\textbf{0.657}&\bf 0.487&0.295&0.129&0.513&0.471\\\hline
        \bf KG-RoBERTa-large&0.324&0.375&0.428&0.307&0.522&0.580&\textbf{0.567}&0.443\\
        \bf KG-T5-large&0.351&0.406&0.457&0.512&0.477&0.419&\textbf{0.567}&0.458\\\hline
        \bf QA-T5-large&0.540&0.468&0.514&0.153&0.613&0.548&0.459&0.471\\\hline
        \bf Retrieval-T5-large &\textbf{0.567}&0.750&0.400&0.205&\bf 0.681&\textbf{0.612}&\textbf{0.567}&\textbf{0.537}\\\hline
	\end{tabular}
\end{table*}

\section{Results}
\label{sec:result}

\subsection{Can Zero-Shot Models Reason over Traffic?}
Table~\ref{tab: all result} shows the overall results for four different types of models (NLI-based, KG-based, QA-based, and retrieval-augmented) on our three benchmarks (BDD-QA, TV-QA, HDT-QA), against a random baseline, supervised, unsupervised model, and human evaluation. Overall, we note that the best performance of our models is in the range between 44 and 54 \% accuracy, which is much better than random and vanilla language model~(Roberta-unsupervised) performance, underscoring the important role that knowledge injection of language models plays. However, it remains significantly behind the supervised models (whose accuracy is between 82 and 90\%) and human performance (accuracy of 95\% or higher). The performance is particularly low on the HDT-QA dataset, which pertains to the intricate details of driving policies and regulations required to reason over this task. 
Among the methods, we note that the most effective approach is to augment QA methods with information retrieved on the fly, which achieves the best performance on two out of three datasets. Namely, on both BDD-QA and HDT-QA, combining DPR with Unified-QA improves the performance of merely using Unified-QA by 2.6 - 6.6\%, which indicates that domain knowledge is essential for decision-making and human driving tests. On the HDT-QA dataset, the NLI and KG models perform comparably to the vanilla LM baseline, implying that general and commonsense knowledge is much less relevant for solving human driving tests. Meanwhile, the best performance on the TV-QA task is achieved by the KG-augmented model. The importance of commonsense knowledge for reasoning over complex and hypothetical situations is intuitive. Curiously, on TV-QA, retrieving domain information is counter-productive and decreases the performance by 2.4\%, possibly highlighting a distraction brought by the domain knowledge for more abstract reasoning tasks.

\noindent 
\subsection{What is the Accuracy per Traffic Action?}
\label{ssec:bddqaresults}
Table \ref{tab: bdd result} shows the fine-grained model results on BDD-QA for different actions.
Although the zero-shot models achieve similar overall performance, there are large differences in their accuracy between action classes. The KG-based models show balanced performance across the 7 classes, while the NLI- and QA-based models exhibit notable differences in performance between action classes. The overall best-performing model, DPR + Unified-QA, is better than the other models for 4/7 classes. However, the NLI-based Roberta-large model performs best on the "Move", ``Slow'', and "Stop" action classes, but poorly in the "Merge" and "Turn" classes. A possible reason for this lies in the nature of the actions like "Move" and "Slow", which are about following the actions of other cars, and can be easier for the NLI-based model to capture due to its strong semantic coherence capabilities. 
Learning from in-domain data also has the lowest impact on the classes Merge and Turn, which may signify that these classes are either more diverse or have less data compared to the others.
However, the KG-based Roberta-large model, which performs worse than the NLI-based model overall, demonstrates better performance on "Merge" and "Turn" actions. We believe that this is inherently linked to the fact that these actions are less probable to be found in NLI corpora, whereas KGs are able to provide richer semantic signals for the representation of merging and turning, an observation consistent with our general findings earler in this section. Finally, we note that both QA method families (with and without information retrieval) perform poorly on the ``Stop'' class, which is also the most difficult for the baseline Roberta-large model. We expect that this indicates that situations that require a vehicle to stop are most difficult, however, further investigation is needed to understand the specific reasons for this observation. 

\begin{table*}[!ht]
\centering
\small
\caption{Results by reasoning type on TV-QA. Random accuracy varies due to different numbers of candidate answers. 
}
\label{tab: tv result}
\begin{tabular}{l | rrrr | r}
\hline
\multirow{2}{*}{\bf Methods}& \multicolumn{5}{c}{\bf Reasoning Task}\\
&Event Forecasting&Reverse Reasoning& Counterfactual~~& ~~Introspective~~~~~& ~~~~~Average\\\hline
\bf Random baseline&0.353&0.299&0.438&0.468&0.373\\
\bf Roberta-large (unsupervised)&0.323&0.382&0.384&\bf 0.571&0.412\\\hline
\bf NLI-Roberta-large&0.535~&0.482~&\bf 0.589~&0.366~&0.479~\\ \hline
\bf KG-Roberta-large&0.457~&0.446~&0.358~&0.535~&0.465~\\
\bf KG-T5-large&\textbf{0.577~}&0.460~&0.384~&0.517~&\textbf{0.506~}\\\hline
\bf QA-T5-large&0.485~&\textbf{0.517~}&0.410~&0.401~&0.468~\\\hline
\bf Retrieval-T5-large&0.464~&0.446~&0.538~&0.383~&0.444~\\\hline
\end{tabular}
\end{table*}

\subsection{What is the Accuracy per Reasoning Type?}
\label{ssec:tvqaresults}

The model accuracy per reasoning task in TV-QA is shown in \autoref{tab: tv result}. Overall, the best-performing model varies per reasoning task. Among the four methods, the KG-augmented models perform the best in forecasting events and introspection. 
The NLI-based model emerges as the clearly best performer on the \textit{Counterfactual} task with a margin of 10\% over the second-best model. This may be due to the fact that \textit{Counterfactual} questions ask about \textit{What could happen if...}, which have a similar pattern to the objectives of the NLI training.
As expected (cf. \autoref{ssec:tv-qa}), we observe that zero-shot reasoning models bring relatively higher improvement over the random baseline on the \textit{Event Forecasting} and \textit{Reverse Reasoning} tasks. 
This can be attributed to the fact that these two tasks rely on information that is already given in the event description (\textit{what will happen next? what might happen before?}), while the other two tasks of Counterfactual and Introspective inference require reasoning over hypothetical events that rely on a deeper understanding of causal relationships and a large amount of background knowledge.
Curiously, while we hypothesize that commonsense knowledge is crucial for complex reasoning, we observe that all of the models perform worse than the Vanilla-Roberta-large model on the \textit{Introspective} task portion, showing that all of the tested models are not capable of answering questions like~\textit{What should have been done to x}.
This indicates the need for alternative modeling architectures that can natively perform multi-hop reasoning.
 

\subsection{Can We Transfer Knowledge to Novel Tasks?}

A key premise of our work is that heterogeneous knowledge about the domain and about event causality should be combined to perform robust reasoning about traffic situations. To investigate this hypothesis further,
we use HDT-QA as a training source and assume a certain amount of traffic domain knowledge can be injected after a model training on HDT-QA. Given the multiple-choice question format of HDT-QA makes it easier for knowledge injection, we use the same training strategy as prior works did~\cite{Ma2021} on knowledge injection: given a multiple-choice QA dataset, the model uses a Masked language Model to give the score of each candidate, then use the Margin Loss Function to maximize the score of the correct answer.
To assess the impact of traffic domain knowledge (HDT) on BDD-QA and TV-QA, we evaluated the HDT-QA trained model on both BDD-QA and TV-QA. We also include commonsense knowledge from the CSKG training dataset, and fuse two training data together. Four experiments are conducted as a result: 1) vanilla Roberta-large, 2) Roberta-large trained on HDT, 3) Roberta-large trained on CSKG, and 4) Roberta-large trained on a fusion of HDT and CSKG dataset.

\begin{figure}[!t]
\centering
\includegraphics[width=\linewidth]{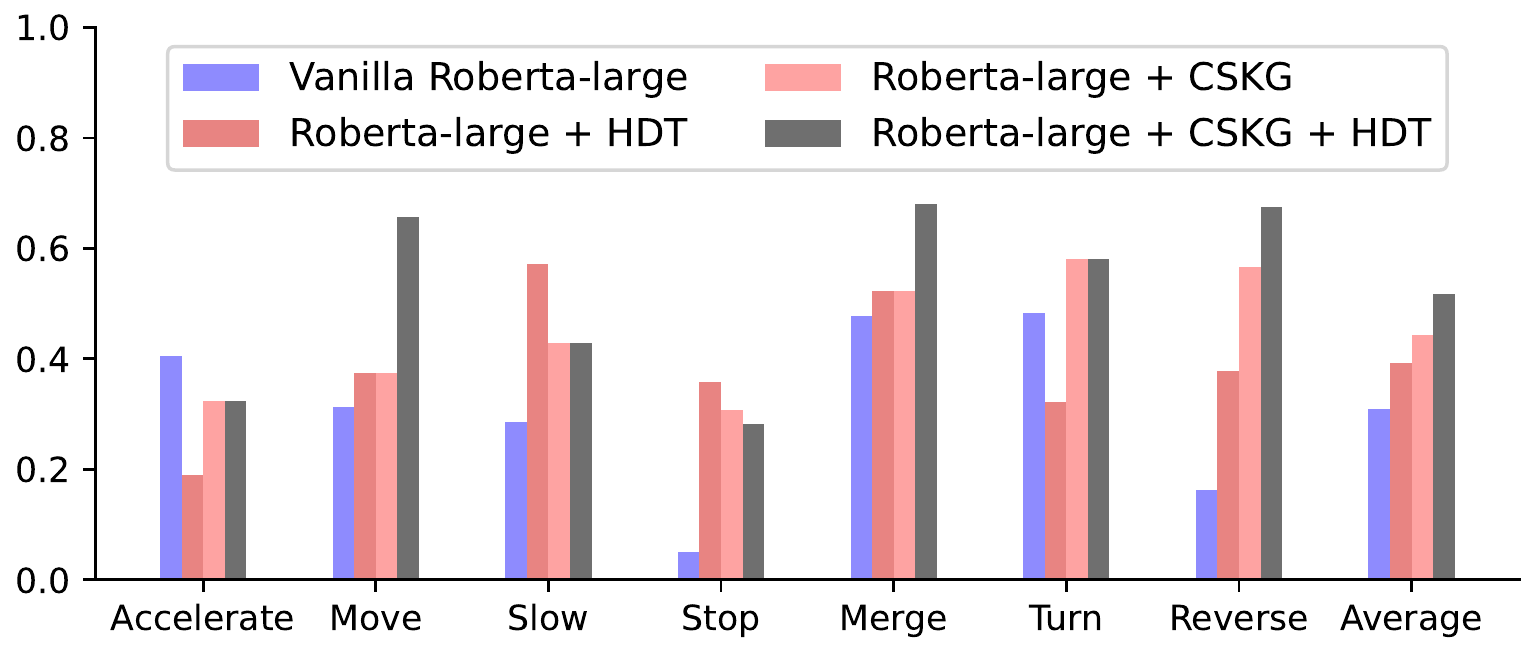}\\
\caption{Granular accuracy of transfer-learning and knowledge fusion on BDD-QA for four model variants per class and on average: 1) vanilla Roberta-large, 2) Roberta-large trained on HDT, 3) Roberta-large trained on CSKG, and 4) Roberta-large trained on a fusion of HDT and CSKG dataset.}
\label{fig: fusion1}
\end{figure}

Figure~\ref{fig: fusion1} show the results of four models' granular performance on the BDD-QA dataset. Injecting common knowledge and driving knowledge is both helpful for making correct traffic decisions, and merging the knowledge leads to better performance. Namely, the $Roberta + CSKG + HDT$ model reaches the best performance on average, followed by $Roberta + CSKG$ and $Roberta + HDT$.
The results show that combining commonsense and domain knowledge results in better performance compared to using only one or the other. The model trained with both knowledge sources indirectly transfers driving knowledge to the knowledge required for answering BDD-QA tasks, while the model trained with only driving knowledge has difficulty answering BDD-QA questions. This is intuitive, e.g., commonsense knowledge can tell language models that cars will pass the crosswalk when driving and crosswalk will appear at the intersection, while traffic domain knowledge tells the models that cars should yield to pedestrians passing the crosswalk. When facing the question~\textit{what might be the reason that a car is waiting in the intersection when the traffic light is green}, the answer~\textit{The car is waiting for pedestrians} will come out, if both knowledge sources are effectively injected into the language model. Meanwhile, we note that the impact of the two knowledge types is not always positive, which confirms our findings in \autoref{ssec:bddqaresults}.

\subsection{Do Model Predictions Overlap?}

\begin{figure}[!t]
\centering
\includegraphics[width=\linewidth]{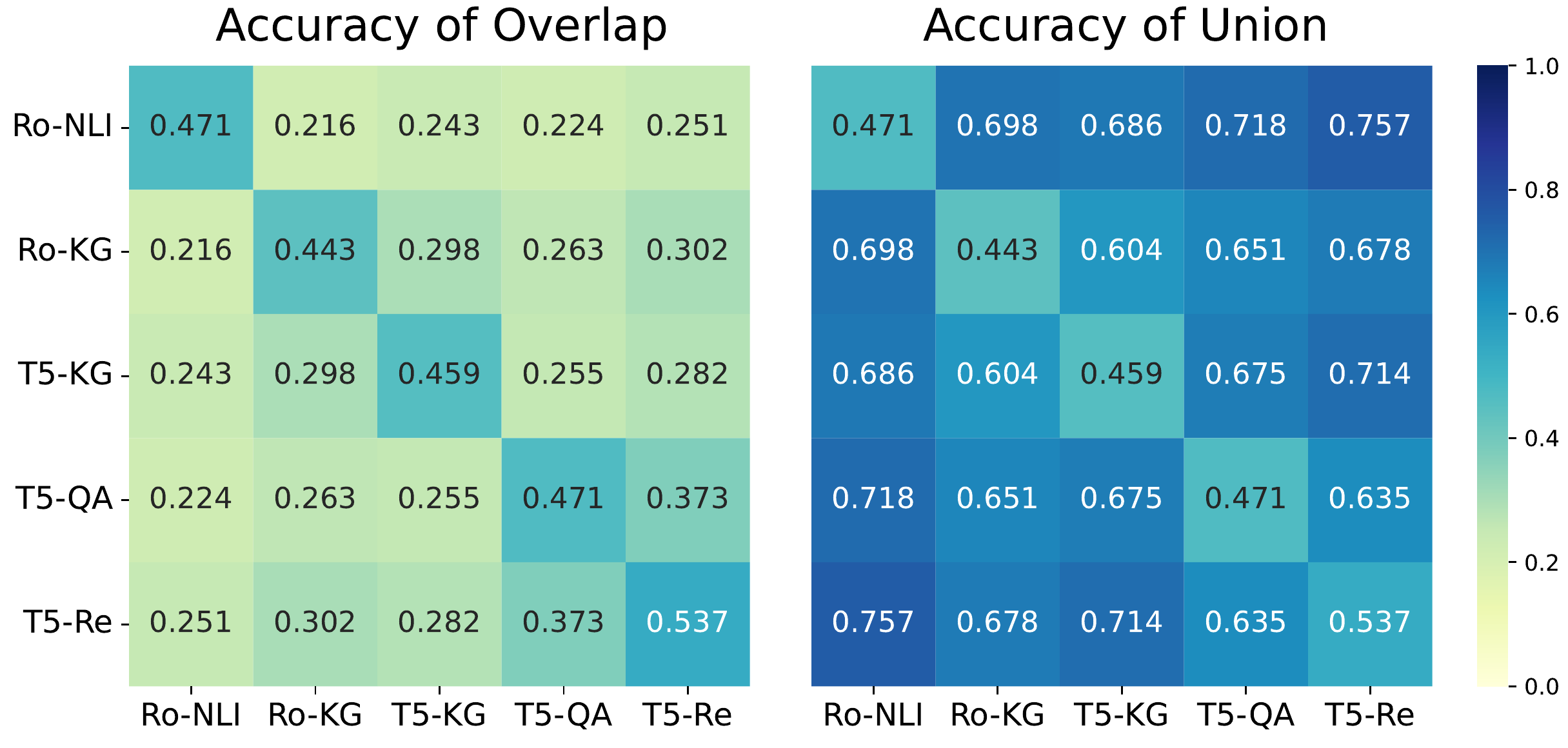}\\
\caption{A heat map representing the intersection (left) and union (right) of correct predictions on BDD-QA for five large models: NLI-Roberta, KG-Roberta, KG-T5, QA-T5, and Retrieval-T5. Each value in the heat map represents 
the joint accuracy of the intersection/union of two models.}
\label{fig: bddoverlap}
\end{figure}

The performance of a model heavily depends on its training data and architecture. To evaluate the predictions of different models, we closely examine the results of five models: \textit{Roberta-large-NLI}, \textit{Roberta-large-KG}, \textit{T5-large-KG}, \textit{T5-large-QA}, \textit{T5-large-IR} on the BDD-QA dataset. We measure the intersection and the union of their accurate predictions, to understand their agreement and the potentially complementary results.
As shown in the left part of Figure~\ref{fig: bddoverlap}, although the models attain similar accuracy, their predictions have limited overlap with each other, with only around 50\% of their correct predictions agreeing. Even among the models \textit{Roberta-large-KG} and \textit{T5-large-KG}, which have the same knowledge source, only two-thirds of their correct predictions overlap. This finding is consistent with the results in~\ref{ssec:bddqaresults} and~\ref{ssec:tvqaresults} that the granular performance of different models is much more diverse than their overall performance. Similarly, we note that the union of the model predictions results in notably higher accuracy (Figure~\ref{fig: bddoverlap} - right). 
Specifically, combining \textit{Roberta-large-NLI} and \textit{T5-large-IR} results in an accuracy of $0.757$, with a gap of only $0.137$ compared to supervised learning methods. These results are much higher than the performance of each model individually and come close to the results obtained from supervised learning, thus motivating the need for the development of methods that combine model predictions and knowledge in a fine-grained manner.

\section{Discussion}

\subsection{Summary of Findings}
In this section, we revisit our research questions and give a brief summary of the findings.

\noindent \textbf{Can knowledge-augmented models reason effectively over traffic situations?}~~Overall, the best knowledge-augmented models reach an accuracy that is much higher than random performance, but the gap between the supervised model's and human's accuracy still remain. 

\noindent \textbf{Can models predict various traffic actions with similar accuracy?}~Although general knowledge-augmented models reach similar overall performance, they utilize different sources of background knowledge to make decisions from various perspectives. This is evident in their varied performance across different action classes, which can usually be traced back to the knowledge used to train these models.

\noindent \textbf{Are some reasoning skills more difficult for the models than others?}~Language models have demonstrated superior performance in solving surface-level reasoning tasks compared to deeper-level reasoning tasks in the traffic domain. In some complex reasoning tasks, even knowledge-driven models underperform compared to vanilla language models.

\noindent \textbf{Can traffic knowledge be transferred between different reasoning tasks?}~The fusion of commonsense knowledge sources and human driving knowledge in the training process has been shown to enhance the performance of language models, surpassing the results achieved from training on only one type of knowledge.

\noindent \textbf{Do methods’ predictions agree with each other?}~Granular performance of different models is much more diverse than their individual performance. Combining the predictions of these models can close the gap with supervised learning.


\subsection{Future Work}

\textbf{Comprehensive access to traffic knowledge} Our experiments have shown the relation between different tasks and model knowledge, and have demonstrated that combining domain and commonsense knowledge is usually beneficial for downstream task performance.
Moreover, we observed that the fine-grained model predictions are often complementary, and their ideal combination would significantly outperform either of the individual models. 
We anticipate that a robust AI model should be able to reason over a comprehensive set knowledge types simultaneously and associate questions with reasoning types, e.g., with methods like mixture-of-experts~\cite{gururangan2021demix}. To facilitate explicit guidance for models on different knowledge types, it is important to systematically delineate the relevant knowledge dimensions and reasoning types, analogous to prior such work in general-domain commonsense reasoning~\cite{ilievski2021dimensions}.

\noindent \textbf{From surface QA to hypothetical reasoning} Our experiments have shown the ability of language models to employ various knowledge to reason on QA tasks. However, pure language modeling mechanisms may not suffice for hypothetical reasoning tasks, such as introspection and counterfactual reasoning. For such tasks, it is important to enhance the model architecture with the ability to develop chains of reasoning that abstract over the provided information and consider the causality of possible outcomes. While it is an open question of how to best perform such reasoning, promising avenues to explore are employing chain-of-thought~\cite{wei2022chain}, self-rationalization techniques~\cite{marasovic2021few}, and graph-based methods~\cite{yasunaga2021qa}.


\noindent \textbf{Generalizing traffic QA to other modalities and domains} Our study in this paper involved a range of QA benchmarks designed to measure the possession of various reasoning and knowledge by representative language models on the traffic domain. The annotations that support the BDD-QA and TV-QA benchmarks are based on videos, which opens the possibility for tighter integration of visual and textual modalities, and for introducing new tasks. We envision using these annotations to evaluate semantic scene search~\cite{chowdhury2021towards} based on factual descriptions, e.g., \textit{give me all scenes where three cars collide}, or even hypothetical situations, e.g., \textit{situations where the accident could have been prevented by more lanes on the freeway.} Similarly, the annotations could be used to evaluate captioning models like SwinBERT~\cite{lin2022swinbert} for the traffic domain. Such benchmarks that represent more tasks and modalities will be crucial in advancing toward the development of practical, neuro-symbolic traffic models. 
Furthermore, expanding our research to additional real-world domains holds significant potential. For example, situated commonsense reasoning under real-world scenarios~\cite{PossibleStories,star} is still challenging for language models.
More practically, in robotics, we can assess the proficiency of language models in interpreting complex instructions and predicting mechanical failures~\cite{saycan}. Similarly, other domains like healthcare, finance, and scientific research could benefit greatly from language models' advancements in situational reasoning.


\noindent \textbf{Investigating the capabilities of large language models}
Our research has extensively examined the inference and comprehension abilities of accessible language models, i.e., models with less than 10 billion parameters, in a variety of traffic scenarios. However, the remarkable comprehension and generalization capabilities of larger language models - those with typically more than 10 billion parameters, like GPT-4 \cite{gpt4} — suggest an immense potential for situational reasoning within complex fields. While we do not expect that these models will perform robust situational reasoning out of the box, the prospect of harnessing their sophisticated reasoning abilities to produce intricate insights and forecasts in the dynamic domain of traffic presents an intriguing avenue for future research.

\section{Conclusions}

This paper devised a study framework that investigates the ability of knowledge-enhanced language models to exhibit generalizable reasoning over traffic situations. The framework included three novel text-based benchmarks for multiple-choice question answering: BDD-QA, which evaluates driver-centric decision-making; TV-QA, which evaluates four complex reasoning tasks; and HDT-QA, which measures the possession of domain knowledge. We adopted four popular methods based on natural language inference, knowledge graph self-supervision, question answering, and retrieval-augmented question answering. Our zero-shot experiments revealed that the models are able to reason over these situations to some extent, and the variance in their performance is largely tied to the correspondence between the model knowledge and the task properties. Retrieval-augmented QA models performed best on BDD-QA and HDT-QA, while commonsense knowledge-based models performed best on the more complex reasoning tasks. Further experiments showed that model and knowledge combinations can enhance the robustness of these models. Future work should investigate the comprehensive organization of traffic knowledge, the development of methods that can perform more complex reasoning, and the development of broader suites of evaluation tasks beyond question answering.

\clearpage

\bibliographystyle{ACM-Reference-Format}
\balance
\bibliography{sample-base}

\appendix
\clearpage
\section{What is the Effect of Scaling up Model Sizes?}
To understand the effect of scaling up the model size, we test 3 different sizes of Unified-QA models (T5-large, T5-3B, T5-11B) on our datasets. We focus our analysis on the benchmarks BDD-QA and TV-QA. 

\begin{figure}[!t] 
\centering
\includegraphics[width=\linewidth]{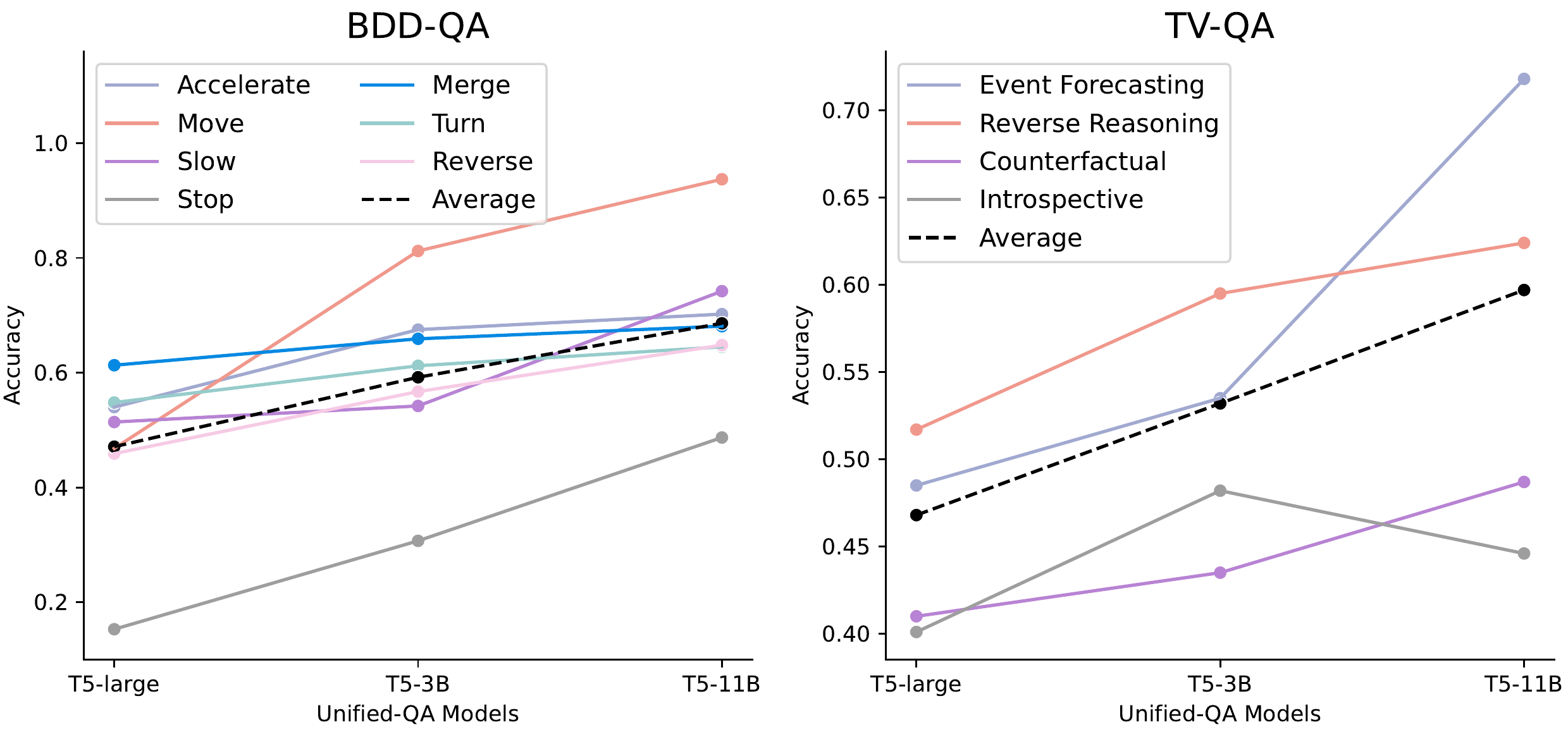}\\
\caption{The result of 3 different sizes of Unified-QA model tested on BDD-QA and TV-QA datasets, the Average result is emphasized using dotted lines.}
\label{fig: size}
\end{figure}

\para{BDD-QA}~~Figure~\ref{fig: size} - left shows the relationship between model size and performance for various traffic actions. In general, an increase in model size leads to improved performance. However, the magnitude of improvement varies among action classes. For instance, the "Move" class exhibits the most substantial improvement, rising from $0.468$ to $0.937$ as the model size grows. Conversely, the "Merge" class exhibits a comparatively modest improvement of $0.068$ points. It is worth mentioning that, despite having one-thirtieth the number of parameters, the Roberta-large NLI-based model achieves the same accuracy in the "Stop" class as the largest Unified-QA model, T5-11B. These results highlight the substantial influence of diverse knowledge sources on the model performance.

\begin{figure}[!t] 
\centering
\includegraphics[width=\linewidth]{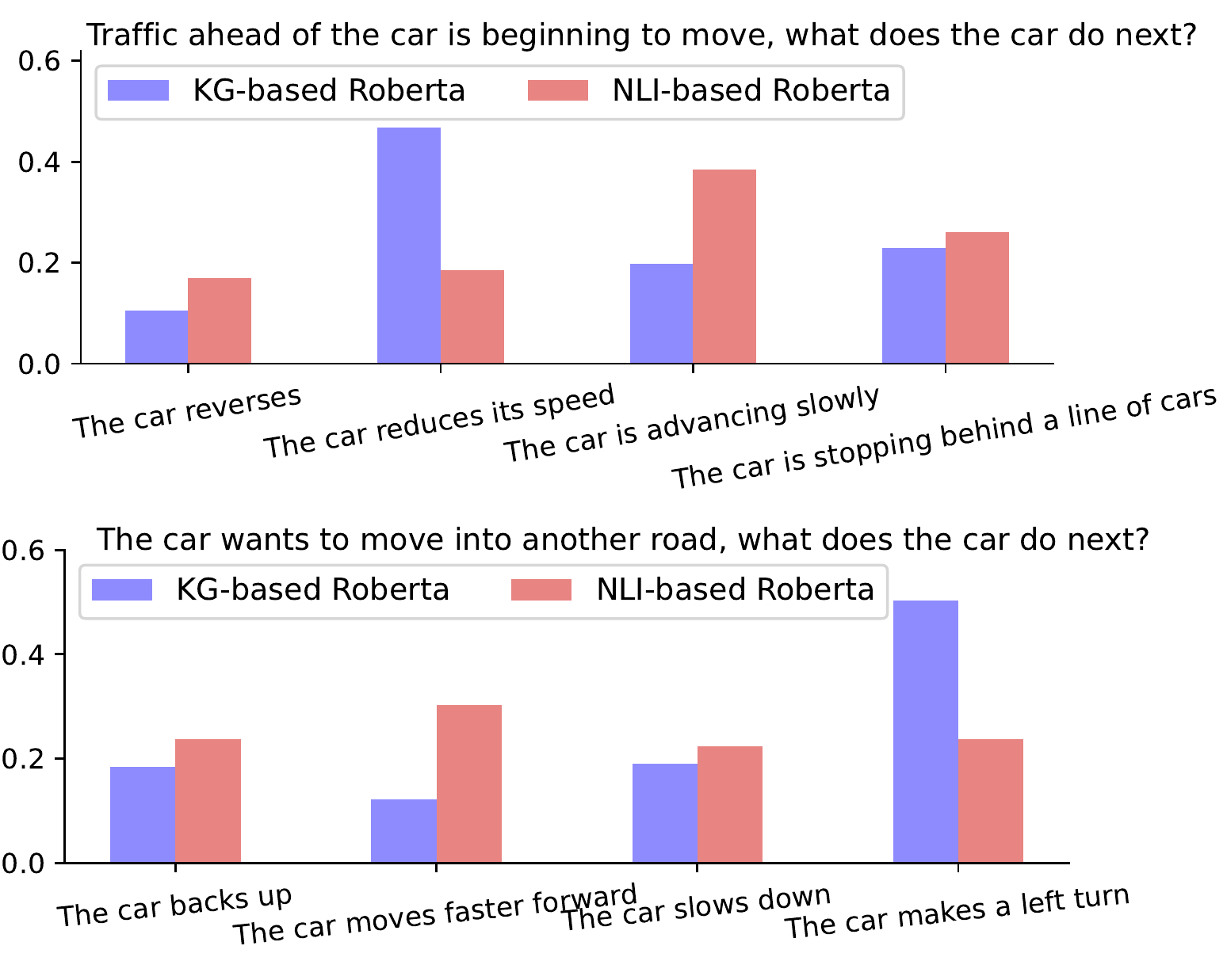}\\
\caption{Prediction distributions for KG- and NLI-based Roberta for two exemplar BDD-QA questions. Questions are shown on the top, and candidate answers are presented on the x-axis. The y-axis shows the probability scores of each answer outputted by the model through a softmax layer.}
\label{fig: case}
\end{figure}

\para{TV-QA}~~In~\ref{fig: size} - right, we observe that as the model size increases, the performance improvement for surface-level reasoning tasks~(\textit{Event Forecasting} and \textit{Reverse Reasoning}) becomes increasingly evident. However, it's noteworthy that for the \textit{Introspective} task, the model performance actually decreases with increased model size, which deviates from typical observations of other QA tasks. 

\noindent In conclusion, while we find that increasing the model size is typically beneficial, its impact varies across reasoning types and action classes. Moreover, the impact of using different sources of knowledge is often bigger than the impact of increasing the model size.

\section{Case Study on the Role of Knowledge in Model Reasoning}

In section~\ref{ssec:bddqaresults}, we have observed that the BDD-QA NLI-based models are more likely to choose the actions that follow other cars', while KG-based models can capture more abundant relations due to the structure of the knowledge graph. To illustrate this finding, we use two examples of traffic domain decision-making questions and show two models' prediction distributions. From the first plot of~\autoref{fig: case}, given the question \textit{Traffic ahead of the car is beginning to move, what does the car do next?}, the NLI-based model chooses the correct answer \textit{The car is advancing slowly}, while the knowledge model chooses the wrong answer \textit{The car reduces its speed}. As we discussed, the NLI-based model solves this question by capturing the semantic coherence between \textit{move} and \textit{advance}. For the second plot of~\ref{fig: case}, given the question \textit{The car wants to move into another road, what does the car do next?}, the NLI-model still chooses the answer based on semantic similarity without considering the fact that the car is moving to another road. Meanwhile, the KG-based model makes the correct choice by successfully connecting the car action \textit{turn} with the description \textit{move into another road}.

\end{document}